# Single-trial EEG Discrimination between Wrist and Finger Movement Imagery and Execution in a Sensorimotor BCI

A.K. Mohamed, T. Marwala, and L.R. John

*Abstract*—A brain-computer interface (BCI) may be used to control a prosthetic or orthotic hand using neural activity from the brain. The core of this sensorimotor BCI lies in the interpretation of the neural information extracted from electroencephalogram (EEG). It is desired to improve on the interpretation of EEG to allow people with neuromuscular disorders to perform daily activities. This paper investigates the possibility of discriminating between the EEG associated with wrist and finger movements. The EEG was recorded from test subjects as they executed and imagined five essential hand movements using both hands. Independent component analysis (ICA) and time-frequency techniques were used to extract spectral features based on event-related (de)synchronisation (ERD/ERS), while the Bhattacharyya distance (BD) was used for feature reduction. Mahalanobis distance (MD) clustering and artificial neural networks (ANN) were used as classifiers and obtained average accuracies of 65 % and 71 % respectively. This shows that EEG discrimination between wrist and finger movements is possible. The research introduces a new combination of motor tasks to BCI research.

## I. INTRODUCTION

PEOPLE who suffer from motor impairments can benefit greatly from a system that can return some of the essential functionality of the human hand [1]. Such people may have had an arm amputated or have suffered a stroke or spinal cord injury [1]. The lost hand of an amputee can be replaced by a robotic prosthetic hand, while the non-functional hand of a victim of a stroke or spinal cord injury can be supported by a robotic exoskeletal orthotic hand [1]. These external devices can then be controlled using the user's thoughts with the help of a brain-computer interface (BCI) to reroute the signals directly from the brain to actuators in the prosthetic/orthotic hand [1, 2].

This solution can be used to allow motor-impaired individuals to perform essential hand movements that facilitate the performance of daily activities [1, 3]. Considering the movements that patients learn during motor rehabilitation [4, 5], five basic hand movements are considered i.e. wrist extension (WE), wrist flexion (WF), finger extension (FE), finger flexion (FF) and the tripod pinch (TR). Occupational therapists consider these to be the most essential hand movements [4, 5, 6].

The core of an effective BCI solution will require that the neural information associated with the essential hand movements be extracted and translated from neural signals, such as electroencephalogram (EEG), in real-time [7, 8]. The combination of these five essential hand movements has not yet been explored in EEG-based BCI literature [9]. It is thus necessary to first investigate the possibility of interpreting the EEG for the five hand movements offline on a single-trial basis since this serves as a first step toward real-time BCI functionality [1, 9, 10]. BCI literature has shown that discriminating the EEG for different movements in a mutliclass problem and EEG discrimination of movements on the same limb are challenging tasks [9, 11]. However, success has been shown in the classification of binary combinations of four types of wrist movement tasks on the same hand [12, 13]. This suggests that the binary classification of other types of unilateral hand movements may be possible. To date, a study has not been conducted to differentiate between major parts of the hand i.e. the wrist and fingers [9, 12, 13, 14]. Hence, as an intermediate step, the differentiation between EEG for wrist and finger movements is investigated in this paper by grouping WE and WF into one class and FE, FF and the TR into another. This forms part of the effort to improve on the incomplete understanding between central neural signals and hand movements [2, 15].

## II. BACKGROUND

### A. Electroencephalogram and ICA

Electrical potentials originating from multiple sources i.e. neuron clusters, combine to form a superposition of topographical maps on the scalp, which can be measured by scalp electrodes to form EEG [17]. There are several challenges associated with the extraction of relevent information from EEG. The signals are small (in the µV range), difficult to measure and are easily contaminated by artifacts [1]. EEG also presents a large inter-trial and inter-subject variability [1]. The billions of simultaneously-active neural processes are measured from a limited number of EEG electrodes (even with high resolution EEG e.g 128 electrodes) [1, 18]. This results in a considerable mixing of information sources from all over the head at each electrode [1, 17, 19]. However, clinical research has increased the understanding of EEG signals and numerous studies have shown relationships between EEG and imagined movements [1, 8, 20, 21]. Inexpensive computer equipment now supports the required computational demands for EEG signal processing [1]. The latter factors make it possible to use EEG as a signal source for basic prosthetic/orthotic hand control [1] in a controlled laboratory environment.

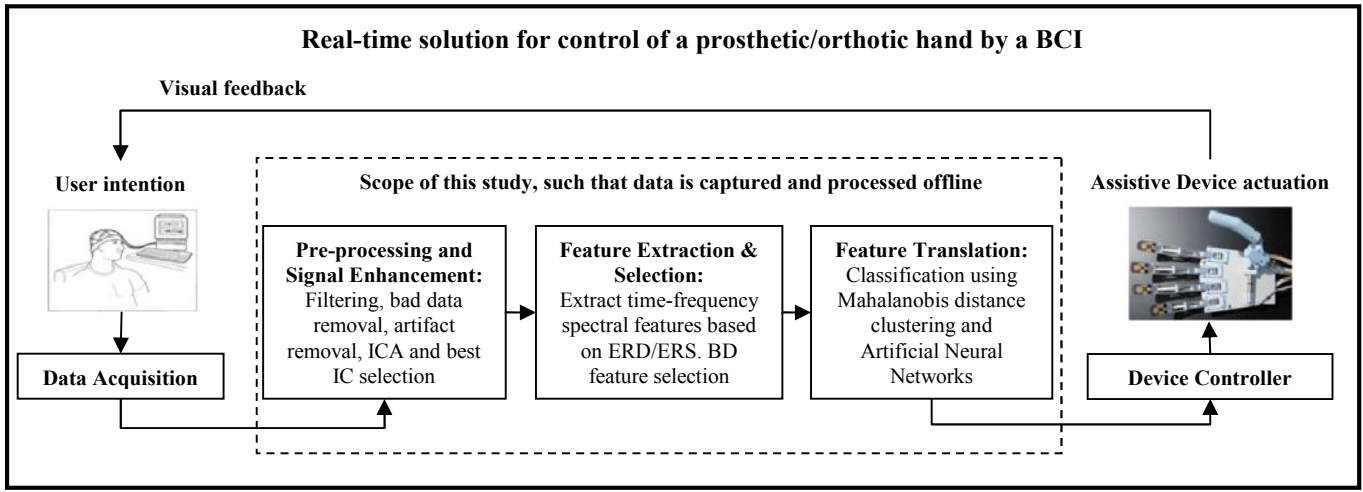

Fig. 1. Model of a sensorimotor BCI used for communication to a prosthetic hand.

Using independent component analysis (ICA), measured signals consisting of a linear mixture of statistically independent source signals, such as EEG, can be decomposed into their fundamental underlying independent components (IC) thus extracting the original source signals [16, 22].

ICA is commonly used in BCI research to remove artifacts, but has also proven useful in separating biologically plausible brain components whose activity patterns relate to behavioural occurrences [18]. In some studies, ICA has shown superior performance over other methods of spatial filtering [23, 24] and has aided the discrimination of EEG for different unilateral wrist movement tasks [13]. This suggests that it may be beneficial for isolating rhythmic activity from the sensorimotor cortex for other types of hand movments [1, 16].

### B. Brain-computer Interface

By using EEG or other electrophysiological methods, a BCI provides a communication channel from the brain to the external world, circumventing the natural neuro-muscular pathway [1, 7]. They can improve the quality of life for those who suffer from motor impairments [1, 25].

The main components of a BCI are shown in Fig 1. They enable execution of the external device according to the user's intent [1, 25]. BCIs that deal with motor functions or sensory inputs of the body deal with the sensorimotor cortex of the brain. They are thus called sensorimotor BCIs and are ideal for the control of a prosthetic/orthotic hand. Prominent electrophysiological features associated with the brain's normal motor output channels are mu (8–12 Hz) and beta (13–30 Hz) rhythms [1, 25]. The rhythms are synchronised when no sensory inputs or motor outputs are being processed [1, 25]. Movement or preparation for movement results in a desynchronisation (decrease in amplitude) of the mu and beta rhythms, referred to as event-related desynchronisation (ERD) [1, 25, 26]. Event-related synchronisation (ERS) occurs after movement when the rhythms synchronise (increase in amplitude) again [1, 25, 26]. ERD and ERS occur during imagined movements as well, making them suitable for paralysed individuals [1, 3]. Features based on ERD/ERS have been used successfully to differentiate the EEG for some types of wrist movements [12, 13].

### III. METHODOLOGY

Fig 1 summarises the major processes that make up the method in order to classify between unilateral wrist and finger movements. The process is applied to real and imagined movements.

### A. Data Acquisition

Subsequent to ethics approval from the University of Cape Town, data was captured from five right-handed, healthy, male, untrained volunteers in their early twenties. The subjects were seated in a comfortable chair, resting their forearm on an arm rest [12, 13]. A computer screen was used along with custom Eprime software [27] to queue the movements while the subjects' EEG were measured. An EGI system that consisted of 128 high-impedance scalp electrodes (forming the GSN 128) along with the Geodesic EEG System and Net Station Software was used [28]. The electrodes were Ag/Ag-Cl electrodes with sponge attachments soaked in an electrolyte solution of potassium chloride [29].

Each subject was asked to perform real and imagined repetitions of the 5 movement sets for each hand (starting with the right hand). Therefore, for each hand, the subjects performed 10 sets of movements: 5 for real movements and 5 for imagined movements. Each set consisted of 20 repetitions/trials of one type of movement [13]. The order of the sets was randomised and thus differed for each subject so that no movement type was preferred [14]. In summary each test subject performed: movement set (5) × L/R hand (2) × real/imagined (2) × repetitions (20) = 400 trials.

The type of movement for each set was shown to the subjects on the computer screen prior to the commencement of the set and a brief practice session was allowed. There were short breaks between sets and the repetitions for each set were performed continually. The trials were queued by instructions shown on the computer screen, the timeline of which is shown in Fig 2 [13, 30].

Subjects were asked not to blink, swallow, move their eyes, adjust their bodies or clear their throats during S1 and S2, but rather during S3, so as to reduce artifact contamination [30]. Any undesired movements or behaviour by the subjects was noted.

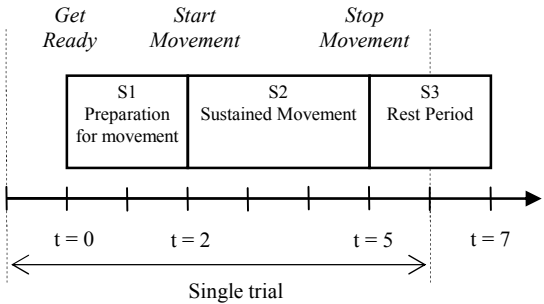

Fig 2. Time sequence and instructions for a single trial

### B. Pre-Processing

EEGLAB was used to handle the pre-processing [18]. Noisy channels were removed and a bandpass filter between 0.5 Hz and 100 Hz was applied to the data [13, 30], which was sampled at 200 Hz by the EGI system [29]. A 50 Hz notch filter was also applied [24].

Data was then divided into 7 s trials, from t = -1s to t = 6 s, placing t = 0 at the *Get Ready* event (pre-movement stimulus) shown in Fig 2. This was done so that the continuous signals were not split in the crucial areas of S1 and S2. Bad trials were removed after manual inspection for voltage spikes and severe distortions across multiple channels. The left hand data for subjects 1 and 4 was unusable and thus discarded.

The Automatic Artifact Removal (AAR) toolbox for EEGLAB [31] was used to remove artifacts, which included electro-oculogram from eye-blinks and eye movements, and electromyogram from tongue, face, neck and shoulder movements [1]. Artifacts were removed using spacial filtering and blind source separation [31]. A bandpass filter between 8 – 30 Hz was then applied to isolate and mu and beta data [12].

### C. ICA and Source Localisation

ICA was run using the infomax algorithm on the individual hands of each subject [18]. This decomposed the EEG into separable localised sources of potentials. The potentials or ICs emanating from the motor cortex were visually selected and isolated.

Several ICs representing motor activity were selected per subject and per hand. This approach is advantageous since the inter-subject variability of EEG makes it difficult to predict which electrodes provide relevant information [32]. It also helps to capture the information from different regions of the motor areas, which may activate during different stages of movement [32]. Furthermore, it reduces the dimensionality of the data and filters contamination from non-sensorimotor neural potentials, such as the visual alpha rhythm [25]. The number of selected ICs varied between test subjects, ranging between 8 and 12. The criteria for selection are based on:
1. Viewing localised activity mainly in the region of the primary motor cortex that controls the hand, but activity in the supplementary motor area and premotor area is also considered [32, 33].
2. The presence of ERD just prior to and/or during S2 as well as ERS after S2 [34]. This is calculated using the inter-trial variance method [34].

### D. Feature Extraction and Selection

A time-frequency technique was used to extract power spectral features from the selected ICs due to the non-stationary nature of EEG [25]. The time range from t = 1 s to t = 4 s was considered (see Fig 2) in order to include pre-movement and movement execution/imagination phases. An overlapping sliding window of 300 ms was then applied in increments of 100 ms [13, 24]. The power spectrum for each window was calculated using an FFT. The frequency spectrum was then split into 7 bands of 3 Hz each [30] and the sum of the powers within each band formed a feature. 28 time windows were extracted over the time range considered, with 7 power band features each. This was done for each IC, resulting in a total number of features ranging between 1568 and 2352.

The Bhattacharyya distance (BD) was used to select the best features according to how well each feature separated the classes [24, 30]. Hence the BD was calculated for each feature and the 18 features with the largest BD were selected. This provided low dimensionality and was found to be the optimum number of features during iterative testing.

### E. Classification

A clustering classifier based on the Mahalanobis distance (MD) is simple and robust and has shown good performance in BCI research [7]. The MD measures the dissimilarity between feature vectors from different classes and can also be used to remove outliers [35]. Multilayer perceptron artificial neural networks are used widely in BCI research [7] and are used to verify and possibly improve on the MD classification results.

The squared MD $d_i^2$ between the $i^{th}$ vector of dataset $x$ and the mean of dataset $y$ can be calculated using (1), where $\mu_Y$ is the mean of dataset $y$ and $C_Y^{-1}$ is the inverse covariance matrix of dataset $y$ [36].

$$d_i^2 = (x_i - \mu_Y)^T C_Y^{-1} (x_i - \mu_Y) \quad (1)$$

The MD is then used to calculate the distance between each trial in a given class to its own mean and to the mean of the other class [36]. If the distance between a single-trial feature vector $x_i$ and the mean of its class $\mu_x$ is smaller than the MD between that single-trial vector and the mean of the other class, then it can be concluded that $x_i$ belongs to class $x$. The trial being tested is removed from the calculations of the means and covariances of the classes/clusters allowing all trials to be used for testing.

Alternatively, for classification using artificial neural networks (ANNs), the data is divided into training and testing data in a 7:3 ratio. The number of hidden nodes is iteratively varied to select that which yields the smallest average error for all subjects. Hence, MLPs each consisting of 18 input nodes, 24 hidden nodes and 1 output node are trained per subject per hand.

In clinical applications, sensitivity and specificity are often used to evaluate the accuracy of diagnostic tests [37]. Sensitivity describes the likelihood of a positive test result if a patient has a disease, while specificity indicates the likelihood of a negative result if the patient does not have the disease [37]. Sensitivity and specificity can be generalized to 2 class datasets, for example: wrist movements = positive test result and finger movements = negative test result. Classification accuracy is thus measured by calculating the average of the sensitivity and specificity measures (*SSA*) as shown in (2), where *T* and *F* respectively represent the number of correctly and falsely classified trials for each class. Subscripts *W* and *F* denote wrist and finger classes respectively.

$$SSA = \frac{1}{2}\left(\frac{T_W}{T_W + F_W} + \frac{T_F}{T_F + F_F}\right) \quad (2)$$

## IV. RESULTS AND DISCUSSION

The MD and ANN results are summarised in Table I and Table II respectively. Classification is shown per subject for real and imaginary movements. The results show reasonable classification accuracies, which are consistent across most test subjects for both hands. ANNs performed better than MD clustering. This is probably due to the ANNs managing to capture the hidden patterns amongst the features more accurately than the simple distance-based approach of the MD method.

Classification is slightly more successful for imagined movements than for real movements. This is contrary to the findings of other BCI studies [30], where classification results for real movements are superior due to real movements generating stronger motor neural activity [30, 39]. However, some studies have shown similar results for real and imagined movements [13]. The superior results for imagined movements in this study could be due to the fact that all the test subjects were university students who were familiar with motor imagery. Consequently their concentration levels and imaginative skills may have been above average, which may have increased the classification accuracy for imagined movements [40]. Subjects who participated in the study in [12] reported an ease of imagining movements such as WE since it is used in everyday life. In this study, the use of WE, WF, FE, FF and the TR in everyday life may have made the motor imagery tasks easier for the test subjects, thus enhancing their sensorimotor EEG patterns, despite having no training.

The success of this research is important since it shows that the discrimination of neural signals from neighbouring areas of the motor cortex is possible using EEG. This allows the real or imagined movement of major parts of the hand i.e. the wrist and fingers, to be interpreted via EEG. The use of ICA along with high resolution EEG (128 channels) played an important role in this regard. Common hand movements such as FE and the TR [4. 5], which are novel to BCI literature, can be explored in future research involving prosthetic/orthotic hand control using a BCI [9].

Future work involves working towards accurately classifying the individual five essential hand movements; first offline and thereafter in real-time.

TABLE I
CLASSIFICATION ACCURACY (%) FOR EEG DISCRIMINATION BETWEEN WRIST AND FINGER MOVEMENTS USING MD CLASSIFIERS

|  | Real | | Imaginary | |
|---|---|---|---|---|
|  | RH | LH | RH | LH |
| **Subject 1** | 68 | - | 61 | - |
| **Subject 2** | 63 | 84 | 56 | 54 |
| **Subject 3** | 62 | 45 | 69 | 63 |
| **Subject 4** | 71 | - | 76 | - |
| **Subject 5** | 49 | 55 | 81 | 70 |
| **Subject Average** | 63 | 61 | 69 | 62 |
| **Grand Average** | 65 % | | | |

TABLE II
CLASSIFICATION ACCURACY (%) FOR EEG DISCRIMINATION BETWEEN WRIST AND FINGER MOVEMENTS USING ANN CLASSIFIERS

|  | Real | | Imaginary | |
|---|---|---|---|---|
|  | RH | LH | RH | LH |
| **Subject 1** | 81 | - | 70 | - |
| **Subject 2** | 73 | 75 | 79 | 68 |
| **Subject 3** | 73 | 56 | 61 | 72 |
| **Subject 4** | 70 | - | 76 | - |
| **Subject 5** | 52 | 69 | 82 | 67 |
| **Subject Average** | 70 | 67 | 73 | 69 |
| **Grand Average** | 71 % | | | |

## V. CONCLUSION

This paper focuses on discriminating between unilateral wrist and finger movements in order to improve EEG interpretation to allow a sensorimotor BCI to control a prosthetic/orthotic hand. The average results for the MD and ANN classifiers are 65 % and 71 % respectively. These results show that the offline discrimination between wrist and finger movement EEG, for real and imagined movements, is possible. This is an important step towards allowing a prosthetic/orthotic hand to perform essential hand movements.